\documentclass{article}
\usepackage{spconf,amsmath,graphicx}

\usepackage{url}
\usepackage[colorlinks,linkcolor=blue, anchorcolor=black,citecolor=black]{hyperref}
\usepackage[small]{caption}
\usepackage{graphicx}
\usepackage{amsmath}
\usepackage{amsthm}
\usepackage{booktabs}
\usepackage{algorithm}
\usepackage{algorithmic}
\usepackage{amssymb} %for \mathbb{R} and so on
\usepackage[switch]{lineno}
\usepackage{color}
\usepackage{subfigure}
\usepackage{booktabs}

\usepackage{multirow}
\usepackage[table,xcdraw]{xcolor}
\usepackage{arydshln}

\title{WHC: Weighted Hybrid Criterion for Filter Pruning on Convolutional Neural Networks}
\name{Shaowu Chen, Weize Sun, Lei Huang
\address{Department of Electronic and Information Engineering, Shenzhen University}
\thanks{Corresponding author: Weize Sun (proton198601@hotmail.com).}
\thanks{
This research was supported by
the Guangdong Basic and Applied Basic Research Foundation
under Grant 2021A1515011706, the National
Science Fund for Distinguished Young Scholars under Grant
61925108, the Foundation of Shenzhen under
Grant JCYJ20190808122005605,   the Open Research Fund from the Guangdong Provincial Key Laboratory of Big Data Computing, The Chinese University of Hong Kong, Shenzhen, under Grant B10120210117-OF07.}}

\begin{document}
\ninept

\maketitle

%######################################################################################################################
\begin{abstract}
Filter pruning has attracted increasing attention in recent years for its capacity in compressing and accelerating  convolutional neural networks.
% To alleviate performance degradation caused by filter pruning,
Various data-independent criteria, including norm-based and relationship-based ones,
were proposed to prune the most unimportant filters.
%  following the ``pre-train\to\toprune\to\tofine-tune'' three-stage pipeline.
However,
these state-of-the-art criteria fail to fully consider the dissimilarity of filters, and thus might lead to performance degradation.
In this paper,
we first analyze the limitation of relationship-based criteria with examples,
and then introduce a new data-independent criterion,  Weighted Hybrid Criterion (WHC), to tackle the problems of both norm-based and relationship-based criteria.
By taking the magnitude of each filter and the linear dependence between filters into consideration,
WHC can robustly recognize the most redundant filters, which can be safely pruned without introducing severe performance degradation to networks.
Extensive pruning experiments in a simple one-shot manner demonstrate the effectiveness of the proposed WHC.
In particular,
WHC can prune ResNet-50 on ImageNet with more than 42\% of floating point operations reduced without any performance loss in top-5 accuracy.
\end{abstract}


%######################################################################################################################
\section{Introduction}\label{Introduction}

% ({\textbf{When take about cos,and 90 degree, remember: 1) point out FPGM prefer 180 in the introduction when illustrating cases; 2) take about 90 degrees, ???(One question here, 90 projection is the best approximation, then is 90 degree a good indicator?--maybe yes, because they are two totally different questions, one is to select filters to represent the whole space, while the other assert that the projection (with certain length in 90 degree space) limited in that 90 space)}})

% {\textbf{CFP}} measure the correlation, near zero would be pruned.
% {\textbf{Meta}} how about cosine?

\begin{figure}[ht]
\centering
\subfigure[Similar norms, different angles]{\label{case1}
\includegraphics[width=0.39\linewidth]{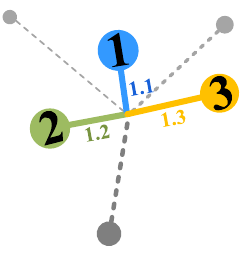}}
\hspace{10mm}
\subfigure[Similar angles, different norms]{\label{case2}
\includegraphics[width=0.39\linewidth]{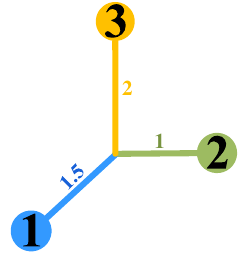}}
%\subfigure[WHC]{\label{WHC_illustration}
%\includegraphics[width=1\linewidth]{WHC}}
\caption{Examples in which relationship-based criteria lose efficacy.}
\label{case}
\end{figure}

Deep convolutional neural networks (CNNs) have achieved great success in various research fields  in recent years,
and broader or deeper architectures have been derived to obtain better performance~\cite{ResNet2016He}.
However,
state-of-the-art CNNs usually come with an enormously large number of parameters,
costing prohibitive memory and computational resources,
and thus are problematic to be deployed on resource-limited platforms such as mobile devices \cite{JointCompression}.
%especially when real-time outputs are required.

To tackle this problem,
pruning methods,
including weight pruning \cite{han2015learning, LotteryTicket} and filter pruning approaches \cite{HYBRIDicassp22, BatchNormLiu2017ICCV,DBLP:conf/icassp/ZhangYW22},
are developed to compress and accelerate CNNs.
Weight pruning evaluates the importance of elements of weight tensors using criteria such as absolute value~\cite{han2015learning},
and sets those with the lowest scores to zero to achieve element-wise sparsity.
Nevertheless,
as the sparsity is unstructured,
weight pruning relies on customized software and hardware to accelerate CNNs.
%the realistic acceleration achieved by weight pruning  is restricted.
%weight pruning relies on customized software and hardware to implement speedup and compression,
%making it less practical.
%By contrast,
%the filter pruning methods~\cite{PFEC,CFP,KDicassp22},
%%on the other hand,
%can produce structured sparsity by removing redundant filters to reduce memory and time consumption of CNNs efficiently on  general-purpose hardware and BLAS libraries,
%%can efficiently save the memory consumption and time for forwarding inference on general-purpose BLAS libraries and hardware by removing redundant filters to  yield slimmer CNNs,
%
%
%can efficiently reduce  memory and inference time consumption on general-purpose BLAS libraries and hardware by removing redundant filters to  yield slimmer CNNs,
%By contrast,
%the filter pruning methods~\cite{PFEC,CFP,KDicassp22} remove structured filters, which yields slimmer CNNs that consume less  memory and inference time on general-purpose BLAS libraries and hardware, and thus has attracted more attention in recent years.
By contrast,  filter pruning methods remove structured filters, which yields slimmer CNNs that can be applied to general-purpose hardware using the common BLAS libraries directly with less memory and inference time consumption, and thus have attracted more attention in recent years.

One of the core tasks in filter pruning is to avoid severe performance degradation on CNNs.
To this end, many pruning criteria, including data-driven \cite{BatchNormYe2018ICLR, molchanov2019importance, yu2022hessian} and data-independent ones \cite{PFEC,KDicassp22},  are proposed to find the most redundant filters, which can be safely deleted.
In this paper, we focus on  data-independent criteria, which can  be further divided into two categories:  norm-based \cite{SFP} and relationship-based \cite{KDicassp22,FPGM,CosineDistance,CFP}.
The former  believe that norms such as the $\ell_1$ and $\ell_2$~\cite{PFEC,SFP}  of filters indicate their importance,
and thus prune filters with  more minor norms.
%Believing that norms such as $\ell_1$~\cite{PFEC} and $\ell_2$~\cite{SFP} norms of filters indicate their importance,
%norm-based criteria proposes to delete filters with smaller norms.
However,
He {\em{et al.}}~\cite{FPGM} argue that it is difficult for norm-based criteria to verify unimportant filters when the variance of norms is insignificant.
%He {\em{et al.}} \cite{FPGM} argue that the small variance  of norms between filters may disable norm-based criteria.
To solve the problem,
they propose a relationship-based criterion, FPGM,
which prunes filters with the shortest Euclidean distances from the others.
In line with FPGM,
cosine criterion~\cite{CosineDistance}  and CFP~\cite{CFP} are also developed, in which filters with the highest angles and the weakest correlation with others are considered the most valuable, respectively.

Generally speaking, relationship-based methods can overcome the problems introduced by norm-based criteria but still have their imperfections.
For example,
considering the colored filters shown in Figure \ref{case1} with similar small norms but different angles,
FPGM and the cosine distance criterion will delete the 1st filter while keeping the inverted 2nd and 3rd since they have the largest Euclidean or angle-wise distance with others.
However,
the 2nd and 3rd filters will extract strongly correlated (although negatively) feature maps that contain highly redundant information,
while the 1st filter orthogonal to them may extract totally different features.
Deleting the 1st filter will weaken the representative capacity of CNNs,
therefore,
it is the 2nd or 3rd filter that should be deleted instead of the 1st one.
Furthermore,
considering that the 2nd filter has a smaller norm than the 3rd one,
it is more reasonable to prune the 2nd filter.
Whereas,
the cosine distance \cite{CosineDistance} and CFP ~\cite{CFP} criteria may rate the 2nd and 3rd the same score and remove one of them randomly.
Similarly,
this situation would also happen when filters have similar angles,
%The similar situation would also happen when filters have  similar angles,
such as the example shown in the Figure \ref{case2}.
These examples demonstrate that relationship-based criteria also need improvement.
%although the variance of their norms are larger enough to tell which ones should be pruned.

% in a 3-D case shown in Figure \ref{},
% cosine distance and LRF are negated when the filters are at similar angles.
% Furthermore,
% in 2-D subspace when pruning three filters with similar norms shown in Figure \ref{},
% when one filter should be removed,
% FPGM and cosine distance will prune the black filter that is orthogonal to others,
% while filters linearly correlated (although negatively) are kept.
% This is not unreasonable,
% since the pruned filter is orthogonal to others and thus can bring a feature map that enhances the representative capacity of CNNs,
% while the kept filters are replaceable by each other.

In this paper,
we propose a Weighted Hybrid Criterion (WHC) that considers both magnitude and relationship of filters
%includes the advantages of both norm-based and relationship-based criteria
to address the problems mentioned above and alleviate performance degradation robustly.
Specifically,
we value filters that have both more significant norms and higher dissimilarity from others (manifesting as orthogonality, instead of antiphase in FPGM \cite{FPGM} and cosine distance criterion \cite{CosineDistance})
%at the same time
 while deleting the others.
Moreover,
we weigh filters' dissimilarity terms differently rather than equally.
That is,
when evaluating a filter,
its dissimilarity terms with filters of more significant norms are assigned a greater weight,
while for filters of more minor norms,
lower weights are appointed.
The reason for this is that the dissimilarity evaluated by the degree of orthogonality is more trustworthy if the norm of a counterpart is more significant.
In this manner,
WHC can rationally score filters and prune those with the lowest scores, {\em{i.e.}}, the most redundant ones,
and thus alleviate the degradation in CNNs' performance caused by pruning.
\section{Methodology}
\subsection{Notation and Symbols}\label{Preliminaries}
%The major notation and symbols are defined as follows:

\textbf{Weight tensors of a CNN.}
Following the conventions of PyTorch,
we assume that a pre-trained $L$-layers CNN has weight tensors $\{\mathcal{W}_l \in \mathbb{R}^{N_{l+1}\times N_{l} \times K \times K} | l=1,2,\cdots,L\}$,
where $\mathcal{W}_l$, $K\times K$, $N_l$ and $N_{l+1}$ stand for the weight tensor of the $l$-th convolutional layer,
the kernel sizes, the number of input and output channels of the $l$-th convolutional layer, respectively.
%(The number of output channels for the $l$-th layer equals to the one of input channels in the ($l$+1)-th layer.)

\textbf{Filters.}
We use $\mathcal{F}_{li}$ to represent the $i$-th filter of the $l$-th layer,
where $\mathcal{F}_{li}=\mathcal{W}_l[i,:,:,:]\in \mathbb{R}^{N_l\times K \times K}$,
{\em{i.e.}},
the $i$-th slide of $\mathcal{W}_l$ along the first dimension.

\textbf{Pruning rates.} $r_{l}=\frac{\#\textrm{pruned filters}}{N_{l+1}} \in [0,1]$ denotes the proportion of pruned filters in the $l$-th layer.

%
%\textbf{Norm matrix.}
%We use ${\textbf{M}}_l$ to represent the diagonal norm matrix of a weight tensor $\mathcal{W}_l$, where
%\begin{equation*}
%    {\textbf{M}}_l =
%    \begin{bmatrix}
%\Vert\mathcal{F}_{l1}\Vert_2 & \ & \ & \ \\
% \ & \Vert\mathcal{F}_{l1}\Vert_2 &\ &\  \\
% \ & \ & \ddots &\  \\
% \ & \ & \ & \Vert\mathcal{F}_{l1}\Vert_2 \\
%\end{bmatrix}.
%\end{equation*}
%
%\textbf{Differentiation matrix.}
%We use ${\textbf{D}}_l$ to represent the differentiation matrix of a weight tensor $\mathcal{W}_l$, where
%\begin{equation*}
%    {\textbf{D}}_l = 1-
%    \begin{bmatrix}
%|\cos{\theta_{1,1}}| & |\cos{\theta_{1,2}}| & \cdots & |\cos{\theta_{1,N_{l+1}}}| \\
%|\cos{\theta_{2,1}}| & \ddots &\ &\vdots  \\
%\vdots& \ & \ddots &\vdots  \\
%|\cos{\theta_{N_{l+1},1}}| & \cdots & \cdots & |\cos{\theta_{N_{l+1},N_{l+1}}}| \\
%\end{bmatrix}.
%\end{equation*}
%The meaning of elements in ${\textbf{D}}_l$ will be introduced in the following content soon.

\subsection{Weighted Hybrid Criterion (WHC)}

Norm-based criteria degrades when the variance of norms of filters is small \cite{FPGM},
and  relationship-based ones may fail to distinguish unimportant filters in several cases, as illustrated in Figure \ref{case1} and \ref{case2}.
To address these problems,
we propose a data-independent Weighted Hybrid Criterion (\textbf{WHC}) to robustly prune the most redundant filters,
which scores the importance of the $i$-th filter $\mathcal{F}_{li}$ in the $l$-th layer by taking into account not only the norm of a filter but also the linear dissimilarity as follows:
\begin{equation}\label{WHC}
{\textrm{score}}_{li} =  \Vert\mathcal{F}_{li}\Vert_2\sum_{j=1, j \neq i}^{N_{l+1}} {\Vert\mathcal{F}_{lj}\Vert_2}\left(1-\lvert\cos{\theta_{i,j}}\rvert\right),
\end{equation}
where
\begin{equation}
\cos{\theta_{i,j}}=\frac{<\mathcal{F}_{li},\mathcal{F}_{lj}>}{\Vert\mathcal{F}_{li}\Vert_2\cdot\Vert\mathcal{F}_{lj}\Vert_2},
\end{equation}
and $\Vert\mathcal{F}_{ij}\Vert_2$ represents the $\ell_2$ norm of the vectorized  $\mathcal{F}_{ij}$.
Note that for a pre-trained model,
we can assume that $\Vert\mathcal{F}_{lj}\Vert_2 > 0$.

When applying WHC in Eq. \ref{WHC} for pruning,
filters with lower scores are regarded as more redundant and thus will be deleted,
while those with the opposite should be retained.
To explain how WHC works in theory,
we first discuss the unweighted variant of (\ref{WHC}), Hybrid Criterion (\textbf{HC}):
\begin{equation}\label{HC}
{\textrm{score}}_{li}' =  \Vert\mathcal{F}_{li}\Vert_2\sum_{j=1}^{N_{l+1}} \big(1-\lvert\cos{\theta_{i,j}}\rvert\big).
\end{equation}
Here $1-\lvert\cos{\theta_{i,j}}\rvert\in \left[0,1\right]$, {the dissimilarity measurement} (\textbf{DM}) between $\mathcal{F}_{li}$ and $\mathcal{F}_{li}$, acts as a scaling factor for $\Vert\mathcal{F}_{li}\Vert_2$,
which in fact widens relative gaps between filters' norms and thus tackles the problem of invalidation of norm-based criteria~\cite{FPGM} caused by a small variance of norms.
Moreover,
unlike euclidean or angle-wise distance-based criteria~\cite{FPGM,CosineDistance} that prefer filters having $180^{\circ}$ angles with others,
WHC (\ref{WHC}) and HC (\ref{HC}) value filters that are more orthogonal to the others,
since they have shorter projected lengths with others and can extract less redundant features.
%and more useful information.

%However,
%the DM term $1-|\cos{\theta_{i,j}}|$ for $j=1,\cdots,N_{l+1}$ in HC (\ref{HC}) are considered equally valuable,
Note that in HC (\ref{HC}),  the DM terms $1-\lvert\cos{\theta_{i,j}}\rvert$ for $j=1,\cdots,N_{l+1}$ are considered equally valuable.
However,
WHC has a different view:
%while we hold the opposite:
when evaluating a filter $\mathcal{F}_{li}$,
weighs for the DM terms should be proportional to  $\Vert\mathcal{F}_{lj}\Vert_2$ for $i \neq j$.
%{\em{i.e.}},
%larger weightings should be appointed when $\Vert\mathcal{F}_{lj}\Vert_2$ are more significant.
The reason for this is that it is less robust to count on filters with smaller norms when scoring filters.
To explain this,
consider two orthogonal filters,
$\mathcal{F}_{l1}=\left(100,0\right)$ and $\mathcal{F}_{l2}=\left(0,0.1\right)$.
Since the norm of $\mathcal{F}_{l2}$ is minor,
 a small additive  interference $\left(-0.1,-0.1\right)$ can easily change $\mathcal{F}_{l2}=(0,0.1)$  to $\mathcal{F}^{'}_{l2}=(-0.1,0)$,
which radically modifies the DM term of $\mathcal{F}_{l1}$ and $\mathcal{F}_{l2}$ from the ceiling 1 to the floor 0.
To improve robustness,
the DM term should be weighted.

\begin{figure}[t]
    \centering
    \setlength{\belowcaptionskip}{-0.3cm} %段后
    \includegraphics[width=0.9\linewidth]{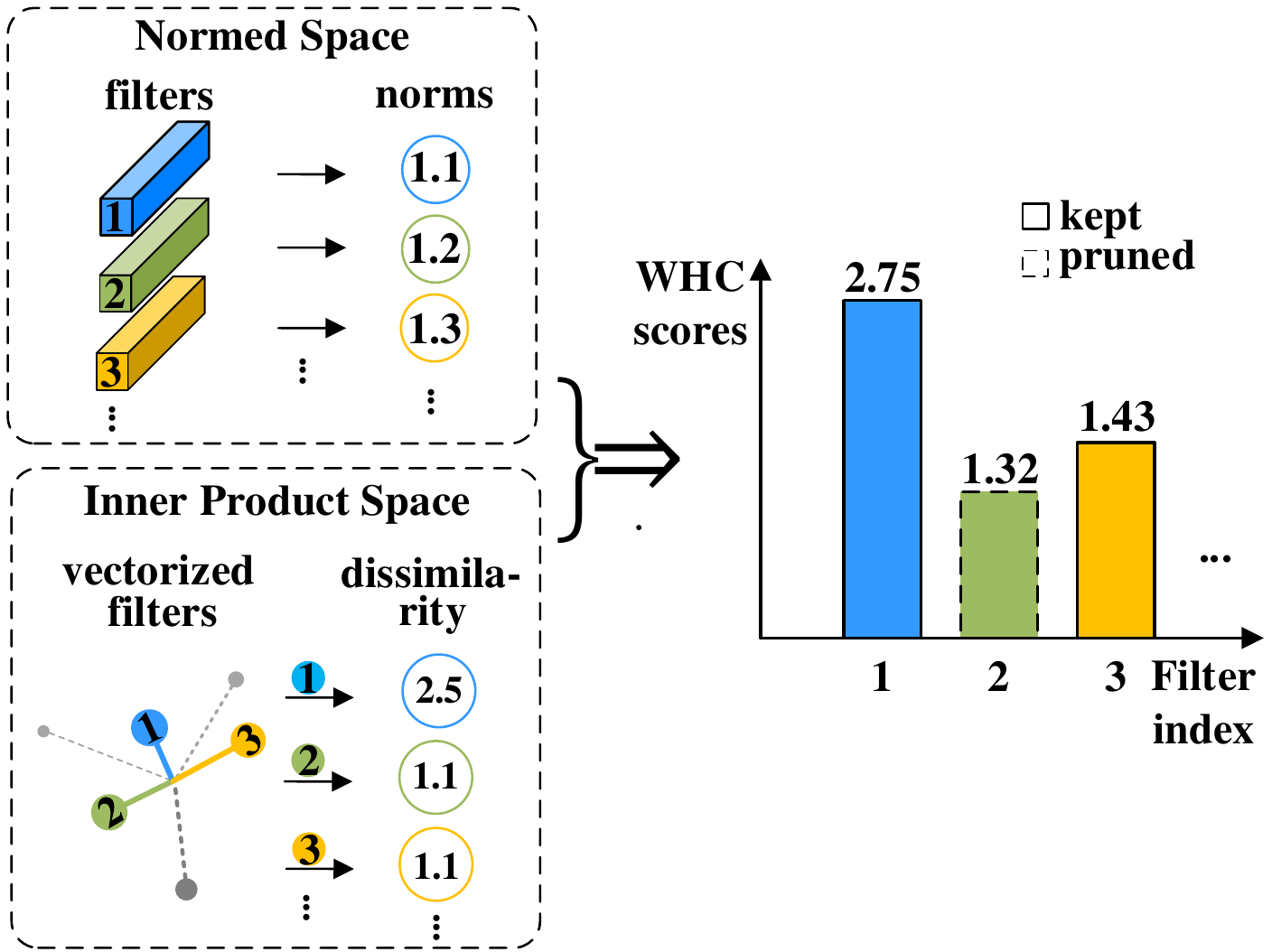}
    \caption{{\em{Left top}}: Magnitude information of filters.
    {\em{Left bottom}}: Measuring dissimilarity of filters.
    {\em{Right}}: WHC scores  filters and prune the most redundant ones.}
    \label{WHCIllustration}
\end{figure}

AutoML techniques such as meta-learning \cite{MetaSurvey22}  can be used to learn the weights,
but it will be time-consuming.
Alternatively,
we directly take norms of filters as weights in WHC (\ref{WHC}),
such that the blind spots of norm-based and relationship-based criteria can be eliminated easily but effectively.
%which is simple but effective to eliminate
As illustrated in Figure \ref{WHCIllustration},
when encountering the case shown in Figure \ref{case1} where norms of filters differ insignificantly,
WHC can utilize dissimilarity information to score the filters and recognize the most redundant one.
Furthermore,
WHC is also robust to the case shown in Figure \ref{case2} in which the filters have similar linear relationships but with varied norms,
while relationship-based criteria such as the cosine criterion \cite{CosineDistance} will lose effectiveness since it scores the filters equally.
There is a scenario that WHC may lose efficacy,
{\em{i.e.}},
filters have the same norms and DM terms at the time.
However,
this indicates that there is no redundancy, and therefore it is not necessary to prune the corresponding model.

\subsection{Algorithm Description}
%{\textbf{(EXPLAIN what single-shot is)}}
As described in Algorithm \ref{algorithm},
we perform filter pruning using WHC following the common ``Pretrain-Prune-Finetune'' pipeline in a simple  single-shot manner, with all layers pruned under the same pruning rate, {\em{i.e.}},
$r_1=r_2=\cdots=r_L$.
Although iterative mechanism \cite{ThiNet}, Knowledge Distillation \cite{Dynamicallyicassp22,ZhouZJMW22,LRF2021Joo}, sensitive analysis that decides layer-wise pruning rates \cite{PFEC}, and some fine-tuning techniques \cite{conf/iclr/LeH21}
%such as
%knowledge distillation
 can improve the performance of pruned CNNs,
none of them are included in this paper for ease of presentation and validation.
% \subsection{Algorithm Description}
% %{\textbf{(EXPLAIN what single-shot is)}}
% As described in Algorithm \ref{algorithm},
% we perform filter pruning using WHC following the common ``Pretrain-Prune-Finetune'' pipeline in a simple  single-shot manner, with all layers pruned under the same pruning rate, {\em{i.e.}},
% $r_1=r_2=\cdots=r_L$.
% Although iterative mechanism \cite{ThiNet}, sensitive analysis that decides layer-wise pruning rates \cite{PFEC}, and some fine-tuning techniques \cite{Dynamicallyicassp22,LRF2021Joo,conf/iclr/LeH21}
% %such as
% %knowledge distillation
%  can improve the performance of pruned CNNs,
% none of them are included in this paper for ease of presentation and validation.
\begin{algorithm}[h]
\caption{WHC for single-shot filter pruning}
\label{algorithm}
\textbf{Input}: Pre-trained model $\{\mathcal{W}_l\}_{l=1}^L$, pruning rates $r_l$, training data,  fine-tuning epoch $epoch_{f}$

\begin{algorithmic}[1] %[1] enables line numbers
\FOR{$l = L \rightarrow 1$}
\STATE Score $\{\mathcal{F}_{li}\}_{i=1}^{N_{l+1}}$ using WHC  (\ref{WHC});
\STATE Prune $r_l*N_{l+1}$ filters with the lowest scores to get $\mathcal{W}'_l$;

\STATE Replace $\mathcal{W}_l$ with $\mathcal{W}'_l$.
\ENDFOR
\STATE Fine-tune $\{\mathcal{W}'_l\}_{l=1}^L$ for $epoch_{f}$ epochs.
%\STATE \textbf{return} $\{\mathcal{W}'_l\}_{l=1}^L$
\end{algorithmic}
\textbf{Output}: Compact model $\{\mathcal{W}'_l\}_{l=1}^L$
\end{algorithm}

By pruning $r_l \cdot N_{l+1}$ filters in the $l$-th layer,
WHC also reduces the same number of input channels in the  $l$+1-th layer.
%Since the output feature maps of the $l$-th layer are the input of the $(l+1)$-th,
%under a pruning rate $r_l$,
%WHC will reduce  $r_l\times N_{l+1}$ filters in the $l$-th layer and the same number of input channels in the  $(l+1)$-th.
%and thus can efficiently reduce floating point operations (FLOPs).
Suppose the input feature maps for the $l$-th layer are of dimensions $H_l \times W_l \times N_l$,
and output feature maps  for the $l$-th, $l$+1-th layer are of dimensions $H_{l+1}\times W_{l+1} \times N_{l+1}$ and $H_{l+2}\times W_{l+2} \times N_{l+2}$,  respectively,
pruning the $l$-th layer with $r_l$ will reduce $H_{l+1}W_{l+1}(N_{l+1}r_l) K^2N_l +H_{l+2}W_{l+2}N_{l+2}K^2( N_{l+1}r_l)$ floating point operations (\textbf{FLOPs}) totally,
which greatly accelerates the forwarding inference.

%#####################################################################################################################

\section{Experiment}

\subsection{Experimental Settings}

\textbf{Datasets and baseline CNNs.}
Following \cite{SFP,FPGM},
we evaluate the proposed WHC on the compact and widely used ResNet-20/32/56/110 for CIFAR-10~\cite{Cifar10Dataset} and ResNet-18/34/50/101 for ILSVRC-2012 (ImageNet)~\cite{ImageNetDataset}
%\footnote{Code and CKPT are available at https://github.com/ShaowuChen/WHC}.
%The reason for pruning these CNNs is that it is more challenging to prune the much less redundant ResNet \cite{ResNet2016He},
%thus the capacity of criteria can be fully demonstrated \cite{PFEC,FPGM}.
For a fair comparison,
we use the same pre-trained models for CIFAR-10 as~\cite{FPGM}.
Whereas, for ILSVRC-2012,
since part of the per-trained parameters of~\cite{FPGM} are not available,
we use official Pytorch pre-trained models~\cite{Pytorch} with slightly lower accuracy.
Code and CKPT are available at \url{https://github.com/ShaowuChen/WHC}.

%We evaluate the proposed WHC on two benchmark datasets: CIFAR-10~\cite{Cifar10Dataset}, and ILSVRC-2012 (ImageNet).
%CIFAR-10 contains 60K $32\times32$ RGB images in 10 categories, including 50K training images and 10K testing ones.
%ILSVRC-2012 is a more extensive dataset containing 1.28 million training images and 50K validation ones in 1K categories.

%\textbf{Baseline CNNs.}
%As stated in \cite{PFEC,FPGM},
%it is more challenging to prune \cite{ResNet2016He}.
%Therefore,
%following \cite{SFP,FPGM},
%we test our WHP on the widely used ResNet-20/32/56/110 for CIFAR-10 and ResNet-18/34/50/101 for ILSVRC-2012.
%For a fair comparison,
%We use the same pre-trained models for CIFAR-10 as ~\cite{FPGM}.
%Whereas for ILSVRC-2012,
%because the trained unpruned baselines of ~\cite{FPGM} are not publicly available,
%we use Pytorch official pre-trained parameters~\cite{Pytorch},
%which have slightly lower accuracy than \cite{SFP,FPGM}.

\textbf{Pruning and fine-tuning.}
The experiments are implemented with Pytorch 1.3.1 \cite{Pytorch}.
We keep all our implementations such as data argumentation strategies, pruning settings, and fine-tuning epochs the same as~\cite{SFP,FPGM},
except that we use the straightforward single-shot mechanism.
In the pruning stage,
all convolutional layers in a network are pruned with the same pruning rate, and we report the proportion of ``FLOPs'' dropped for ease of comparison.

%Different pruning rates $r_l$ are set to achieve reductions on FLOPs  from 28\% to 66\%,
%and all the convolutional layers except projection shortcuts are pruned under the same pruning rate in the pruning stage,
%{\em{i.e.}},
%$r_1=r_2=\cdots=r_L$.
%In accord with \cite{SFP,FPGM},
%after pruning,
%models for CIFAR-10 and ImageNet are fine-tuned for the same epochs as the original baseline CNNs,
%{\em{i.e.}}, 200 and 100 epochs, respectively,
%but with learning rates reduced to one-tenth of the original CNNs.

\textbf{Compared methods.}
We compare WHC with several criteria,
including the data-independent norm-based PFEC \cite{PFEC},  SPF \cite{SFP},  ASPF \cite{ASFP},
relationship-based FPGM \cite{FPGM},
and several data-dependent methods  HRank \cite{HRank}, GAL \cite{GAL},  LFPC \cite{LFPC}, CP \cite{CP},
NISP \cite{NISP},  ThiNet \cite{ThiNet} and ABC \cite{ABC}.

\begin{table}[h]
\setlength{\abovecaptionskip}{0.1cm}  %段前
\setlength{\belowcaptionskip}{-0.3cm} %段后
\scriptsize
% \footnotesize
\centering
\begin{tabular}{ccllcc}
\toprule
\multirow{2}{*}{Depth} &\multirow{2}{*}{Method} &\multirow{2}{*}{\shortstack{ Baseline\\ acc. (\%)}}  &\multirow{2}{*}{\shortstack{Pruned\\acc. (\%)}}   & \multirow{2}{*}{\shortstack{Acc.\\$\downarrow$ (\%)}}    & \multirow{2}{*}{\shortstack{FLOPs\\$\downarrow$ (\%)}}    \\
&&&&\\
\midrule
\multirow{2}{*}{20}    & \textbf{WHC} & 92.20 ($\pm$0.18)  & \textbf{91.62 ($\pm$0.14)}  & \textbf{0.58}    & 42.2  \\
& \textbf{WHC} & 92.20 ($\pm$0.18)  & 90.72 ($\pm$0.16)  & 1.48   & \textbf{54.0}  \\

\midrule

\multirow{2}{*}{32}   & \textbf{WHC} & 92.63 ($\pm$0.70)  &\textbf{92.71 ($\pm$0.08)}  & \textbf{-0.08}    & 41.5  \\
& \textbf{WHC} & 92.63 ($\pm$0.70)  & 92.44 ($\pm$0.12)  & 0.19    & \textbf{53.2}  \\

\midrule

\multirow{15}{*}{56}
& PFEC \cite{PFEC}    & 93.04 & 93.06 & -0.02   & 27.6  \\
& \textbf{WHC}& \textbf{93.59 ($\pm$0.58)} & \textbf{93.91 ($\pm$0.06)} & \textbf{-0.32}   & \textbf{28.4} \\
%\noalign{\vskip 0.2ex}\cdashline{2-6}\noalign{\vskip 0.5ex}
 \cmidrule{2-6}

& GAL \cite{GAL}     & 93.26 & 93.38 & 0.12     & 37.6  \\
& SFP \cite{SFP}    & \textbf{93.59 ($\pm$0.58)}  & 93.78 ($\pm$0.22)  & -0.19    & \textbf{41.1}  \\
& \textbf{WHC}& \textbf{93.59 ($\pm$0.58)} & \textbf{93.80 ($\pm$0.33)} & \textbf{-0.21}   & \textbf{41.1} \\
%\noalign{\vskip 0.2ex}\cdashline{2-6}\noalign{\vskip 0.5ex}
\cmidrule{2-6}
& HRank \cite{HRank}          & 93.26 & 93.17 & 0.09  & 50.0 \\
& SFP \cite{SFP}      & \textbf{93.59 ($\pm$0.58)}  & 93.35 ($\pm$0.31)  & 0.24    & \textbf{52.6}  \\
& ASFP \cite{ASFP} & \textbf{93.59 ($\pm$0.58)}  & 93.12 ($\pm$0.20)  & 0.47   & \textbf{52.6}  \\
& FPGM \cite{FPGM}     & \textbf{93.59 ($\pm$0.58)}  & 93.26 ($\pm$0.03)  & 0.33   & \textbf{52.6}  \\
& \textbf{WHC}& \textbf{93.59 ($\pm$0.58)} & \textbf{93.47 ($\pm$0.18)} & \textbf{0.12} & \textbf{52.6} \\
%\noalign{\vskip 0.2ex}\cdashline{2-6}\noalign{\vskip 0.5ex}
\cmidrule{2-6}
& LFPC \cite{LFPC}    & \textbf{93.59 ($\pm$0.58)}  & 93.24 ($\pm$0.17)  & 0.35  & 52.9  \\
& ABC \cite{ABC}     & 93.26 & 93.23 & 0.03   & 54.1  \\
& \textbf{WHC}& \textbf{93.59 ($\pm$0.58)} & \textbf{93.66 ($\pm$0.19)} & \textbf{-0.07}  & \textbf{54.8} \\
%\noalign{\vskip 0.2ex}\cdashline{2-6}\noalign{\vskip 0.5ex}
\cmidrule{2-6}
& GAL \cite{GAL}    & 93.26 & 91.58 & 1.68   & 60.2  \\
  & \textbf{WHC}& \textbf{93.59 ($\pm$0.58)} & \textbf{93.29 ($\pm$0.11)} & \textbf{0.30} & \textbf{63.2} \\
\midrule

 \multirow{14}{*}{110}
& GAL \cite{GAL}      & 93.50 & 93.59 & -0.09     & 18.7  \\
& PFEC \cite{PFEC} & 93.53 & 93.30 & 0.23  & 38.6  \\
& SFP \cite{SFP}     & \textbf{93.68 ($\pm$0.32)}  & 93.86 ($\pm$0.21)  & -0.18    & \textbf{40.8}  \\
& ASFP \cite{ASFP}   & \textbf{93.68 ($\pm$0.32)}  & 93.37 ($\pm$0.12)  & 0.31   & \textbf{40.8}  \\
& {\textbf{WHC}}& \textbf{93.68 ($\pm$0.32)} & \textbf{94.32 ($\pm$0.17)} & \textbf{-0.64}  & \textbf{40.8} \\
%\noalign{\vskip 0.2ex}\cdashline{2-6}\noalign{\vskip 0.5ex}
\cmidrule{2-6}

& GAL \cite{GAL}      & 93.26 & 92.74 & 0.76    & 48.5  \\
& SFP \cite{SFP}     & \textbf{93.68 ($\pm$0.32)}  & 92.90 ($\pm$0.18)  & 0.78    & \textbf{52.3}  \\
& FPGM \cite{FPGM}     & \textbf{93.68 ($\pm$0.32)}  & 93.74 ($\pm$0.10)  & -0.06    & \textbf{52.3}  \\
& ASFP \cite{ASFP}   & \textbf{93.68 ($\pm$0.32)}  & 93.10 ($\pm$0.06)  & 0.58    & \textbf{52.3}  \\
& {\textbf{WHC}}& \textbf{93.68 ($\pm$0.32)} & \textbf{94.07 ($\pm$0.20)} & \textbf{-0.39}   & \textbf{52.3} \\
%\noalign{\vskip 0.2ex}\cdashline{2-6}\noalign{\vskip 0.5ex}
\cmidrule{2-6}

& HRank \cite{HRank}         & 93.50 & 93.36 & 0.14   & 58.2  \\
& LFPC \cite{LFPC}     & \textbf{93.68 ($\pm$0.32)}  & 93.07 ($\pm$0.15)  & 0.61   & 60.3  \\
& ABC \cite{ABC}      & 93.50 & 93.58 & -0.08     & 65.0  \\
 & \textbf{WHC}& \textbf{93.68 ($\pm$0.32)} & \textbf{93.82 ($\pm$0.07)} & \textbf{-0.14}  & \textbf{65.8}\\
\bottomrule
\end{tabular}
\caption{Pruning results on CIFAR-10. ``$\downarrow$'' means ``drop''. In ``Acc. $\downarrow$'', the smaller, the better; a negative drop means improvement. In ``FLOPs $\downarrow$'', a larger number indicates that more FLOPs are reduced.}
\label{CIFAR10}
\end{table}

%%%%%%%%%%%%%%%%%%%%%%%%%%%%%%%%%%%%%%%
\subsection{Evaluation on CIFAR-10}
For CIFAR-10,
we repeat each experiment three times and report the average accuracy after fine-tuning.
As shown in Table \ref{CIFAR10}, WHC outperforms several state-of-the-art counterparts.
WHC can prune 52.3\%  of FLOPs in ResNet-110 with even 0.39\% improvement,
while the norm-based SFP under the same settings suffers 0.78\% of degradation.
The improvement shows that under moderate pruning rates,
WHC can alleviate the overfitting problem of models without hurting their capacity.

Compared with the iterative ASFP \cite{ASFP}, data-driven HRank \cite{HRank}, AutoML-based ABC \cite{ABC} and LFPC \cite{LFPC},
  WHC in a single-shot manner can also achieve competitive performance.
For example,
although more FLOPs are reduced,
WHC still achieves 0.42\% and 0.75\% higher accuracy than LFPC in ResNet-56 and ResNet-110, respectively,
which demonstrates that WHC can recognize the most redundant filters effectively.
Furthermore,
under similar pruned rates,
as the depth of CNNs increases,
the pruned models obtained by WHC suffer less performance degradation. 
%WHC obtain pruned models with better performance as the depth of CNNs increases.
The reason for this is that deeper CNNs contain more redundancy,
which can be removed by WHC robustly without hurting CNNs' capacity severely.
 %less performance degradation.

\begin{table}[h]
\setlength{\abovecaptionskip}{0.1cm}  %段前
\setlength{\belowcaptionskip}{-0.3cm} %段后
%\footnotesize
\scriptsize

\centering
\tabcolsep=1.8pt
\begin{tabular}{ccccccccc}
\toprule
\multirow{3}{*}{Depth}	   & \multirow{3}{*}{Method}   &\multirow{3}{*}{\shortstack {Baseline\\top-1\\acc. (\%)} }  &\multirow{3}{*}{\shortstack {Pruned\\top-1\\acc. (\%)} }  &\multirow{3}{*}{\shortstack {Top-1 \\acc. $\downarrow$ \\(\%)} } &\multirow{3}{*}{\shortstack{Baseline\\top-5\\acc. (\%)} }   &\multirow{3}{*}{\shortstack {Pruned\\top-5\\acc. (\%)} }    &\multirow{3}{*}{\shortstack {Top-5\\acc. $\downarrow$ \\(\%)} } &\multirow{3}{*}{\shortstack {FLOPs\\ $\downarrow$ (\%)}}   \\
& & & & & & & & \\
& & & & & & & & \\
% \\
% Depth                & Method              & Top1           & AccTop1        & Drop Top1     & Top5           & AccTop5        & Drop Top5     & FLOPS\_drop   \\
\midrule
\multirow{4}{*}{18}    & SFP \cite{SFP}                & 70.23          & 60.79          & 9.44          & 89.51          & 83.11          & 6.40          & \textbf{41.8} \\
                     & ASFP \cite{ASFP}               & 70.23          & 68.02          & 2.21          & 89.51          & 88.19          & 1.32          & \textbf{41.8} \\
                     & FPGM \cite{FPGM}               & \textbf{70.28} & 68.41          & 1.87          & \textbf{89.63} & 88.48          & 1.15          & \textbf{41.8} \\
                     & \textbf{WHC } & 69.76          & \textbf{68.48} & \textbf{1.28} & 89.08          & \textbf{88.52} & \textbf{0.56} & \textbf{41.8} \\
\midrule
\multirow{6}{*}{34}  & PFEC \cite{PFEC}                & 73.23          & 72.17          & 1.06          & -              & -              & -             & 24.2          \\
                     & ABC \cite{ABC}                 & 73.28          & 70.98          & 2.30          & 91.45          & 90.05          & 1.40          & 41.0          \\
                     & SFP \cite{SFP}                 & \textbf{73.92} & 72.29          & 1.63          & \textbf{91.62} & 90.90          & 0.72          & \textbf{41.1} \\
                     & ASFP \cite{ASFP}                & \textbf{73.92} & 72.53          & 1.39          & \textbf{91.62} & 91.04          & 0.58          & \textbf{41.1} \\
                     & FPGM \cite{FPGM}                & \textbf{73.92} & 72.54          & 1.38          & \textbf{91.62} & 91.13          & 0.49          & \textbf{41.1} \\
                     & \textbf{WHC } & 73.31          & \textbf{72.92} & \textbf{0.40} & 91.42          & \textbf{91.14} & \textbf{0.28} & \textbf{41.1} \\
\midrule
\multirow{17}{*}{50} & ThiNet \cite{ThiNet}             & 72.88          & 72.04          & 0.84          & 91.14          & 90.67          & 0.47          & 36.7          \\
                     & SFP  \cite{SFP}               & \textbf{76.15} & 62.14          & 14.01         & \textbf{92.87} & 84.60          & 8.27          & 41.8          \\
                     & ASFP \cite{ASFP}               & \textbf{76.15} & 75.53          & 0.62          & \textbf{92.87} & 92.73          & 0.14          & 41.8          \\
                     & FPGM \cite{FPGM}                 & \textbf{76.15} & 75.59          & 0.56          & \textbf{92.87} & 92.63          & 0.24          & \textbf{42.2} \\
                     & \textbf{WHC } & 76.13          & \textbf{76.06} & \textbf{0.07} & 92.86          & \textbf{92.86} & \textbf{0.00} & \textbf{42.2} \\

                     \cmidrule{2-9}
                     & HRank \cite{HRank}              & \textbf{76.15} & 74.98          & 1.17          & \textbf{92.87} & 92.33          & 0.54          & 43.8          \\
                     & NISP \cite{NISP}               & \textbf{-}     & -              & 0.89          & -              & -              & -             & 44.0          \\
                     & GAL \cite{GAL}                 & \textbf{76.15} & 71.95          & 4.20          & \textbf{92.87} & 90.94          & 1.93          & 43.0          \\
                     & CFP \cite{CFP}                  & 75.30          & 73.40          & 1.90          & 92.20          & 91.40          & 0.80          & 49.6          \\
                     & CP \cite{CP}                  & -              & -              & -             & 92.20          & 90.80          & 1.40          & 50.0          \\
                     & FPGM \cite{FPGM}                & \textbf{76.15} & 74.83          & 1.32          & \textbf{92.87} & 92.32          & 0.55          & \textbf{53.5} \\
                     & \textbf{WHC } & 76.13          & \textbf{75.33} & \textbf{0.80} & 92.86          & \textbf{92.52} & \textbf{0.34} & \textbf{53.5} \\

                     \cmidrule{2-9}
                     & GAL \cite{GAL}                 & \textbf{76.15} & 71.80          & {4.35} & 92.87          & {90.82} & 2.05 & 55.0          \\
                     & ABC \cite{ABC}                  & 76.01          & 73.86          & 2.15          & \textbf{92.96} & 91.69          & 1.27          & 54.3          \\
                     & ABC \cite{ABC}                & 76.01          & 73.52          & 2.49          & \textbf{92.96} & 91.51          & 1.45          & 56.6          \\
                     & LFPC \cite{LFPC}                & \textbf{76.15} & 74.46          & 1.69          & 92.87          & 92.04          & 0.83          & 60.8          \\
                     & \textbf{WHC } & 76.13          & \textbf{74.64} & \textbf{1.49} & 92.86          & \textbf{92.16} & \textbf{0.70} & \textbf{60.9} \\
\midrule
\multirow{4}{*}{101}
% & SFP \cite{SFP}                & \textbf{77.37} & 77.51          & -0.14         & \textbf{93.56} & 93.71          & -0.20         & \textbf{42.2} \\
                     & FPGM \cite{FPGM}               & \textbf{77.37} & 77.32          & 0.05          & \textbf{93.56} & 93.56          & 0.00          & \textbf{42.2} \\
                     & \textbf{WHC } & \textbf{77.37} & \textbf{77.75} & \textbf{-0.38} & 93.55 & \textbf{93.84} & \textbf{-0.30} & \textbf{42.2} \\

                     \cmidrule{2-9}
                     & ABC \cite{ABC}                 & \textbf{77.38} & 75.82          & 1.56          & \textbf{93.59} & 92.74          & 0.85          & 59.8          \\
                     & \textbf{WHC } & 77.37          & \textbf{76.63} & \textbf{0.74} & 93.55          & \textbf{93.30} & \textbf{0.25} & \textbf{60.8}\\
\bottomrule
\end{tabular}
\caption{Pruning results on ILSVRC-2012 (ImageNet). ``acc.'' and ``$\downarrow$'' stand for `'accuracy'' and ``drop'', respectively.}
\label{IMAGENET}
\end{table} 
%%%%%%%%%%%%%%%%%%%%%%%%%%%%%%%%%%%%%%%
\subsection{Evaluation on ILSVRC-2012}\label{eoi}
The results are shown in Table \ref{IMAGENET}.
Not surprisingly,
compared with several state-of-the-art methods,
WHC not only achieves the highest top-1 and top-5 accuracy, but also suffers the slightest performance degradation.
In ResNet-50,
WHC reduces more than 40\% of FLOPs but barely brings loss in the top-1 and top-5 accuracy,
while the norm-based SFP suffers 14\% degradation in the top-1 accuracy and other methods more than 0.5\%.
%%In ResNet-101,
%%the  phenomenon observed in CIFAR-10 also exist,
%%{\em{i.e.}},
%%WHC can even gain 0.38\% improvement in top-1 accuracy, although 42.2\% FLOPs are pruned.
Compared with norm-based and relationship-based criteria,
the superior performance of WHC can be attributed to the utilization of both norm and linear similarity information of filters,
and the assigned weights on different DM terms can provide more robust results.

\subsection{Ablation Study}
\textbf{Decoupling experiment.}
To further validate the effectiveness of WHC,
we progressively decouple WHC into several criteria, as shown in Table \ref{Decoupling}.
The cosine criterion \cite{CosineDistance} is also added for comparison.
% as a baseline.
We repeat pruning 40\% of filters in ResNet-32 and ResNet-56 three times
and report the raw accuracy (without fine-tuning) and the average drop in accuracy after fine-tuning.
Compared with the cosine criterion \cite{CosineDistance},
the DM criterion suffers less degradation in accuracy and is therefore more rational.
Taking into account both norm and dissimilarity,
HC achieves better performance than $\ell_2$ and DM.
Furthermore,
by assigning different weights to the DM terms,
WHC outperforms all counterparts stably,
especially in the more compact ResNet-32.
The $\ell_2$ achieves similar performance as WHC in ResNet-56,
but fails to maintain the same performance in ResNet-32,
demonstrating the robustness of the proposed WHC.
%which indicates norm-based criteria are not as robust as WHC,
%as shown in Subsection \ref{eoi} as well.

% {\renewcommand{\arraystretch}{1.5}
\begin{table}[!h]
\setlength{\abovecaptionskip}{0.1cm}  %段前
\setlength{\belowcaptionskip}{-0.5cm} %段后
%\footnotesize
\scriptsize
% \footnotesize
% \linespread{1.5}
\tabcolsep1.3pt
\centering
\begin{tabular}{cccc}
\toprule
% \multirow{2}{*}{\shortstack{Depth \& \\acc. (\%)}} &\multirow{2}{*}{Criterion}  &\multirow{2}{*}{\shortstack{Acc. $\downarrow$\\(\%)}}\\
% &&\\
\multirow{2}{*}{\shortstack{Depth\& \\acc.}} &\multirow{2}{*}{Criterion}   &\multirow{2}{*}{\shortstack{Raw acc. \\ (\%)}}  &\multirow{2}{*}{\shortstack{Fine-tuned\\acc. $\downarrow$ (\%)}}\\
&&&\\
\midrule
\multirow{5}{*}{\shortstack{32\\\\92.63\%}}
&${\Vert{\mathcal{F}}_{li}\Vert_2}$ \cite{SFP} &10.06  &0.42 \\\noalign{\vskip 0.5ex}
&${  \sum_{j=1}^{N_{l+1}}\cos{\theta_{i,j}}}$~\cite{CosineDistance} &10.00 & 0.78 \\\noalign{\vskip 0.5ex}
&${ \sum_{j=1}^{N_{l+1}}(1-\lvert\cos{\theta_{i,j}}\rvert)}$ (DM)&11.30 & 0.56 \\\noalign{\vskip 0.5ex}
&${\Vert\mathcal{F}_{li}\Vert_2 \sum_{j=1}^{N_{l+1}}(1-\lvert\cos{\theta_{i,j}}\rvert)}$ (HC)&11.18 & 0.32  \\\noalign{\vskip 0.5ex}
&${\Vert\mathcal{F}_{li}\Vert_2 \sum_{j=1}^{N_{l+1}}\Vert\mathcal{F}_{lj}\Vert_2(1-\lvert\cos{\theta_{i,j}}\rvert)}$ (WHC)&11.25& \textbf{0.19}\\
\midrule

\multirow{5}{*}{\shortstack{56\\\\93.59\%}}
&${\Vert{\mathcal{F}}_{li}\Vert_2}$ \cite{SFP} & 16.99 &0.20 \\\noalign{\vskip 0.5ex}
&${  \sum_{j=1}^{N_{l+1}}\cos{\theta_{i,j}}}$~\cite{CosineDistance} &10.00& 0.51 \\\noalign{\vskip 0.5ex}
&${ \sum_{j=1}^{N_{l+1}}(1-\lvert\cos{\theta_{i,j}}\rvert)}$ (DM)&9.94&0.37  \\\noalign{\vskip 0.5ex}
&${\Vert\mathcal{F}_{li}\Vert_2 \sum_{j=1}^{N_{l+1}}(1-\lvert\cos{\theta_{i,j}}\rvert)}$ (HC)&17.94&0.14  \\\noalign{\vskip 0.5ex}
&${\Vert\mathcal{F}_{li}\Vert_2\sum_{j=1}^{N_{l+1}}\Vert\mathcal{F}_{lj}\Vert_2(1-\lvert\cos{\theta_{i,j}}\rvert)}$ (WHC) &19.73&\textbf{0.12} \\
\bottomrule
\end{tabular}
\caption{Decoupling results on ResNet-32 and ResNet-56 for CIFAR-10.  ``acc.'' and  ``$\downarrow$'' stand for ``accuracy''  and ``drop", respectively}
\label{Decoupling}
\end{table}
% } 

%\begin{figure}[ht]
%    \centering
%    \includegraphics[width=0.65\linewidth]{Resnet56.pdf}
%    \caption{Accuracy of ResNet-56 for CIFAR-10 with different proportions of FLOPs pruned.}
%    \label{rates}
%\end{figure}
%
%\textbf{Varying Pruning Rates.}
%To comprehensively reveal the capacity of WHC,
%we change the FLOPs reduced rates by pruning 10\%--90\% of filters in  ResNet-56 for CIFAR-10.
%As shown in Table \ref{rates},
%when the FLOPs prune rate is lower than 40\%,
%the pruned models can even outperform the baseline,
%which demonstrates the capacity of WHC in alleviating the over-fitting problem.
%Furthermore,
%WHC can prune more than 60\% of FLOPs without bringing significant loss.
%
%
\textbf{Types of norm and dissimilarity measurement.}
We replace $\ell_2$ norm and $\cos{\theta_{i,j}}$ in WHC (\ref{WHC}) with $\ell_1$ norm and correlation coefficient, respectively.
The correlation can be regarded as the centralized version of $\cos{\theta_{i,j}}$.
We conduct experiments on ResNet-32 with $r_l=40\%$,
%in which the accuracy of the pruned model produced by the naive WHC is 92.44\pm0.1292.44\pm0.12.
The fine-tuned accuracy of the $\ell_1$ and correlation version of WHC are $(92.50\pm0.11)\%$ and $(92.62\pm0.18)\%$, respectively,
which are slightly higher than the naive WHC, $(92.44\pm0.12)\%$.
The results indicate that WHC can be further improved with more suitable types of norm and dissimilarity measurement.

\subsection{Visualization}
\begin{figure}[!ht]
\setlength{\abovecaptionskip}{0.1cm}  %段前
\setlength{\belowcaptionskip}{-0.5cm} %段后
\centering
\subfigure[Input image]{\label{panda}
\includegraphics[width=0.45\linewidth]{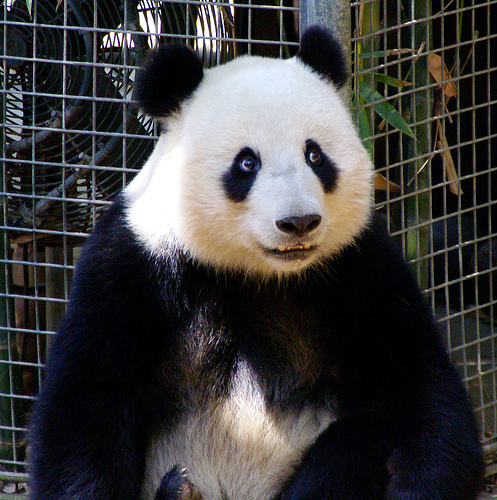}}
\subfigure[Feature maps]{\label{featumre_map}
\includegraphics[width=0.45\linewidth]{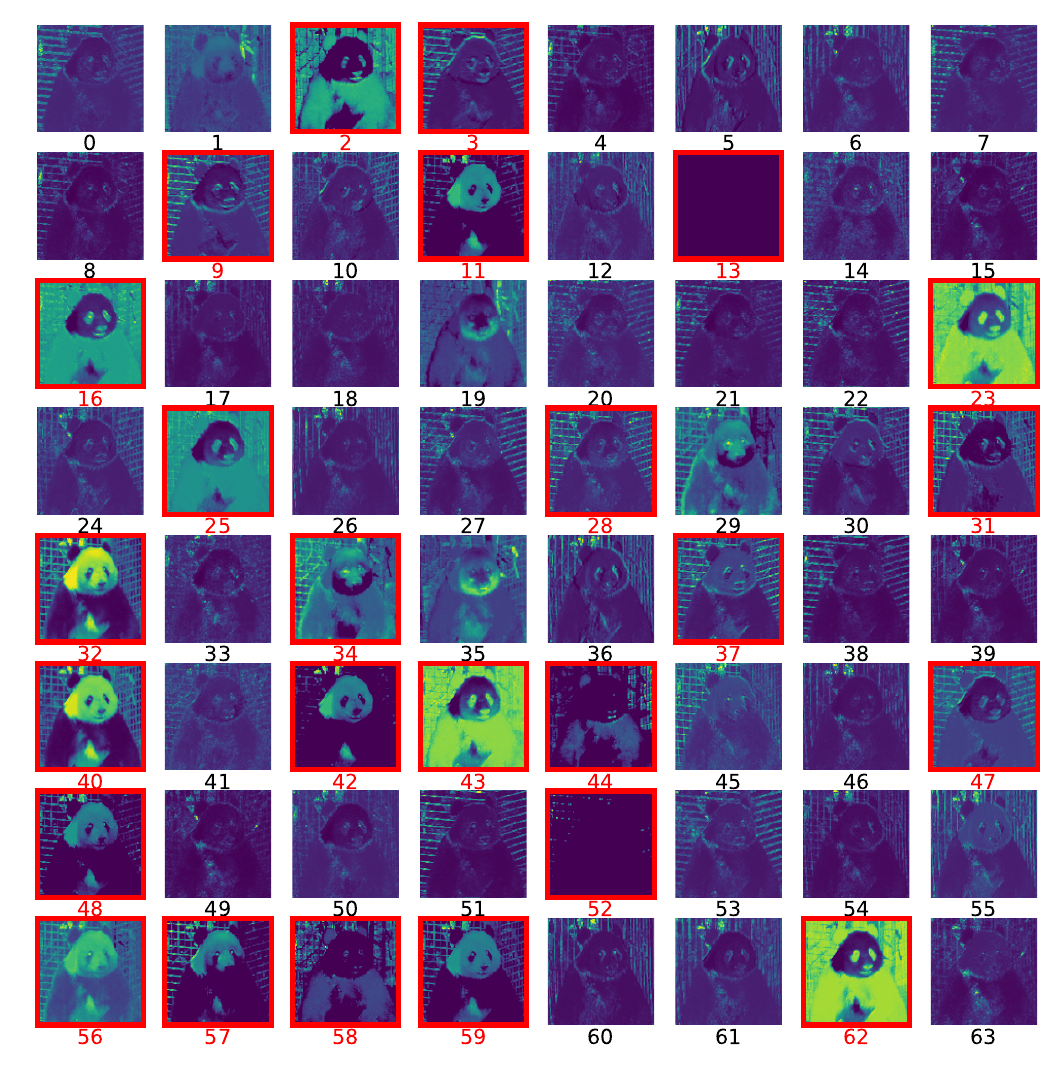}}
\caption{Visualization of ResNet-50-conv1 output feature maps (after ReLU, BN and MaxPooling, {\em{i.e.}},  the input of the second Conv layer). The feature maps bounded by red boxes correspond to the pruned filters; 40\% of the filters are pruned. The highest values are colored in the brightest green, while the lowest in the darkest blue.}
\label{visualization}
\end{figure}

We prune 40\% of filters in the first layer of ResNet-50 for ImageNet and visualize the output feature maps, as shown  in Figure \ref{visualization}.
%Feature maps bounded with red boxes correspond to the pruned filters.
Among the pruned filters,
 3 and 34 can be replaced by 10 and 29, respectively,
and  [13,32,48,62, {\em{et al}}]  fail to extract valuable features.
We also compare WHC with the $\ell_2$ norm criterion \cite{SFP} and relationship-based FPGM \cite{FPGM},
finding that $\ell_2$ and FPGM rank the filters differently from WHC but finally give a similar pruning list under the given pruning rate.
%the pruned list is similar.
The only divergence of WHC and the $\ell_2$ criterion arises over filters 29 and 56:
WHC keeps 29 and prunes 56, while $\ell_2$ criterion takes the opposite action.
We consider filter 29 to be more valuable than filter 56,
since the latter fails to extract significant features,
while the former highlights the eyes of the input image.
The difference of WHC and $\ell_2$ in the pruning list for a single layer is insignificant,
but an accumulation of tens of layers finally results in a wide gap and makes WHC more robust in finding redundant filters.
%different outcomes.

\section{Conclusion}
We propose a simple but effective data-independent criterion,
 Weighted Hybrid Criterion (WHC),
for filter pruning.
Unlike previous norm-based and relationship-based criteria that use a single type of information to rank filters,
WHC takes into consideration both magnitude of filters and dissimilarity between filter pairs,
and thus can recognize the most redundant filters more efficiently.
Furthermore,
by  reweighting the dissimilarity measurements according to the magnitude of counterpart filters adaptively,
WHC is able to  alleviate performance degradation on CNNs caused by pruning robustly.
%Extensive experiments compared with state-of-the-art methods demonstrate the capacity of WHC.

% \section*{Acknowledgments}

%\bibliographystyle{named}
%\bibliography{ijcai_new}
\bibliographystyle{IEEEbib}
\bibliography{cite_new}

\end{document}